\documentclass[letterpaper, 10 pt, journal, twoside]{ieeetran}
\usepackage{amsmath,amsfonts}
\usepackage{algorithm,algpseudocode,algorithmicx}
\usepackage{array}
\usepackage{textcomp}
\usepackage{stfloats}
\usepackage{url}
\usepackage{verbatim}
\usepackage{graphicx, subfigure}
\usepackage{multicol}
\usepackage{multirow}
\usepackage{booktabs}
\usepackage{float}
\usepackage{cite}
\usepackage{siunitx}
\usepackage{tikz}
\usepackage{enumitem}
\usepackage[hidelinks]{hyperref}
\usepackage{colortbl}
\usepackage{threeparttable}

\usepackage{orcidlink}
\usepackage{eso-pic}
\usepackage{everypage}

\newcolumntype{E}{>{\columncolor{lightgray}}c}
\newcolumntype{F}{c}

\newcommand{\sixpt}{\fontsize{6.5pt}{7.2pt}\selectfont}

\AddToShipoutPictureBG*{
  \AtPageUpperLeft{
    \hspace{1.5cm} 
    \raisebox{-\baselineskip}[0pt][0pt]{
      \parbox{\textwidth}{
      \vspace{-0.1cm}
      \centering
        \sixpt{}
        This article has been accepted for publication in IEEE Robotics and Automation Letters. This is the author's version which has not been fully edited and \\
content may change prior to final publication. Citation information: DOI 10.1109/LRA.2024.3460409
      }
    }
  }
}

\AddEverypageHook{%
    \AddToShipoutPictureBG*{
      \AtPageUpperLeft{
        \hspace{1.5cm} 
        \raisebox{-\baselineskip}[0pt][0pt]{
          \parbox{\textwidth}{
          \centering
            \sixpt{}
            This article has been accepted for publication in IEEE Robotics and Automation Letters. This is the author's version which has not been fully edited and \\
    content may change prior to final publication. Citation information: DOI 10.1109/LRA.2024.3460409
          }
        }
      }
    }
}

\AddToShipoutPictureBG*{
  \AtPageLowerLeft{
    \raisebox{\dimexpr 1.5\baselineskip}[0pt][0pt]{
      \parbox{\paperwidth}{ 
        \centering 
        \sixpt{}
        \hspace*{2.25cm}
        \vspace*{-0.5cm}
        © 2024 IEEE. Personal use is permitted, but republication/redistribution requires IEEE permission.
        See https://www.ieee.org/publications/rights/index.html for more information.
        }
    }
  }
}

\AddEverypageHook{%
    \AddToShipoutPictureBG*{
      \AtPageLowerLeft{
      \raisebox{\dimexpr 1.5\baselineskip}[0pt][0pt]{
          \parbox{\paperwidth}{ 
            \centering 
            \sixpt{}
            \hspace*{2.25cm}
            \vspace*{-0.5cm}
            © 2024 IEEE. Personal use is permitted, but republication/redistribution requires IEEE permission.
            See https://www.ieee.org/publications/rights/index.html for more information.
            }
        }
      }
    }
}

\begin{document}

\title{Exact Wavefront Propagation for Globally Optimal One-to-All Path Planning on 2D Cartesian Grids}

\author{Ibrahim Ibrahim\,\orcidlink{0000--0001--6840--558X},
    \and
    Joris Gillis\,\orcidlink{0000--0002--6774--3613}, 
    \and
    Wilm Decré\,\orcidlink{0000--0002--9724--8103},
    \and
    and Jan Swevers\,\orcidlink{0000--0003--2034--5519}

\thanks{Manuscript received: May, 28, 2024; Revised August, 2, 2024; Accepted September, 1, 2024.}
\thanks{This paper was recommended for publication by
Editor M. Ani Hsieh upon evaluation of the Associate Editor and Reviewers’ comments.
This work has been carried out within the framework of Flanders Make's SBO project ARENA (Agile \& REliable NAvigation).} 
\thanks{The authors are with the MECO Research Team, Department of Mechanical Engineering, KU Leuven, Belgium and Flanders Make@KU Leuven, 3000, Belgium. {\tt\footnotesize ibrahim.ibrahim@kuleuven.be}, {\tt\footnotesize joris.gillis@kuleuven.be}, {\tt\footnotesize wilm.decre@kuleuven.be},
{\tt\footnotesize jan.swevers@kuleuven.be}}
\thanks{Digital Object Identifier (DOI): see top of this page.}}

\markboth{IEEE Robotics and Automation Letters. Preprint Version. Accepted September, 2024}
{Ibrahim \MakeLowercase{\textit{et al.}}: Exact Wavefront Propagation for Globally Optimal One-to-All Path Planning on 2D Cartesian Grids} 


\maketitle

\begin{abstract}
  This paper introduces an efficient $\mathcal{O}(n)$ compute and memory complexity algorithm for globally optimal path planning on 2D Cartesian grids. Unlike existing marching methods that rely on approximate discretized solutions to the Eikonal equation, our approach achieves exact wavefront propagation by pivoting the analytic distance function based on visibility. The algorithm leverages a dynamic-programming subroutine to efficiently evaluate visibility queries. Through benchmarking against state-of-the-art any-angle path planners, we demonstrate that our method outperforms existing approaches in both speed and accuracy, particularly in cluttered environments. Notably, our method inherently provides globally optimal paths to all grid points, eliminating the need for additional gradient descent steps per path query. The same capability extends to multiple starting positions. We also provide a greedy version of our algorithm as well as open-source C++ implementation of our solver.
\end{abstract}

\begin{IEEEkeywords}
  Motion and Path Planning, Path Planning for Multiple Robots or Agents, Computational Geometry
\end{IEEEkeywords}

\section{Introduction}

\begin{figure}[ht!]
    \centering
    \includegraphics[width=8cm,keepaspectratio]{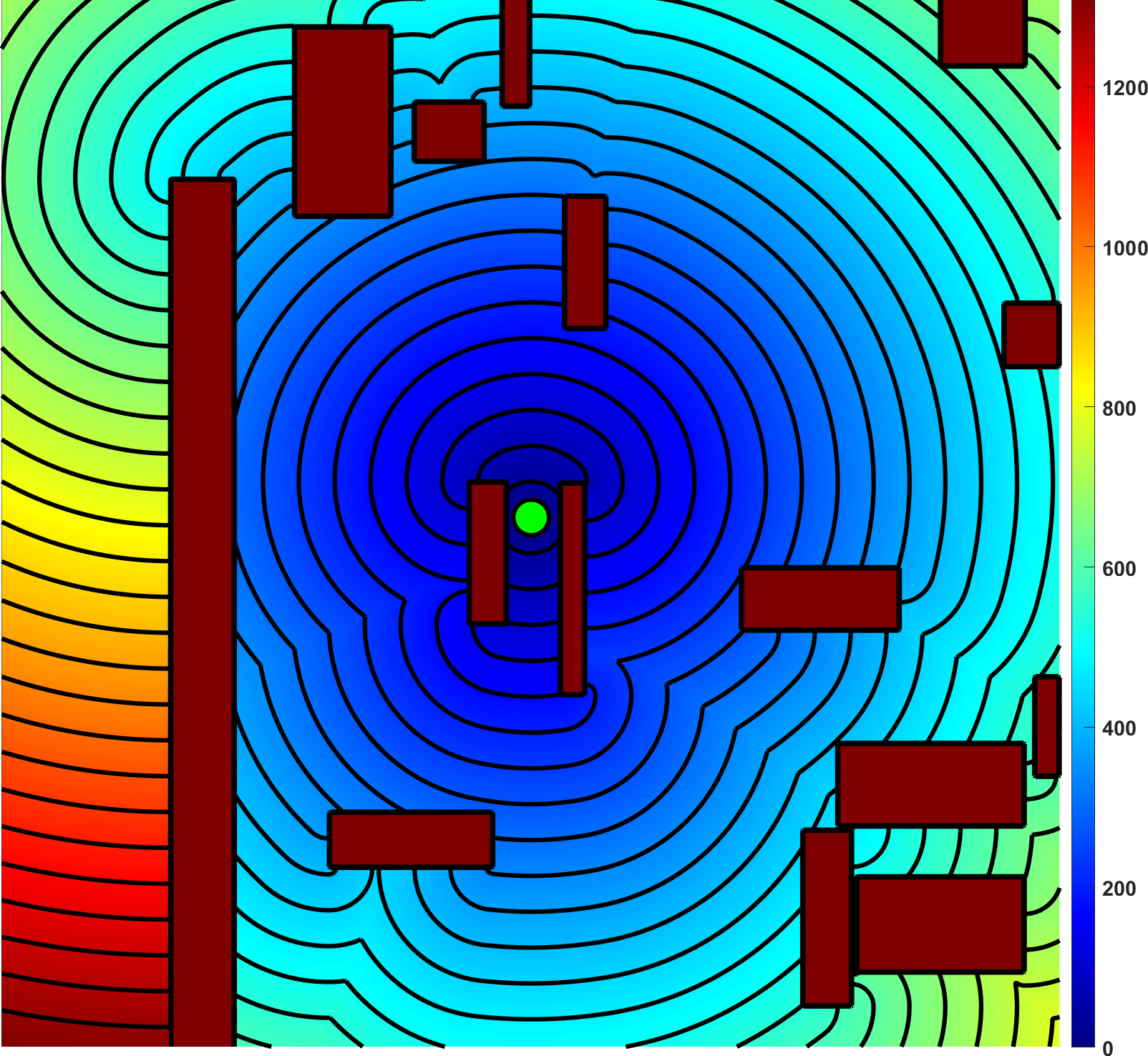}
    \caption{Euclidean distance function starting from the green point at the center. Contour lines are shown in black and the colormap ranges from cold to hot reflecting the increasing distance. Obstacles are shown in dark red. Computed using our proposed method. The colorbar is shown to the right with the distance unit being pixels. The environment is a $1000\times1000$ map. The visualization strategy between grid points is interpolation.}\label{fig:contours_1}
\end{figure}

\IEEEPARstart{W}{avefront} propagation is a powerful numerical method used for solving wave propagation problems in several fields, including computational geometry, fluid mechanics, computer vision, and materials science~\cite{sethian1999level}. In the context of robotics, wavefront propagation has emerged as a promising approach for path planning in obstacle-rich environments. The fundamental principle of this approach is to propagate the wavefront of a distance function (ex. Euclidean), or implicitly a time-of-arrival function, from a starting point to all points in a grid while avoiding collisions with obstacles by wrapping the wavefront around them, e.g., Fig.~\ref{fig:contours_1}. Due to the monotonicity and optimality of the wavefront, the shortest path between the starting point and any other unoccupied point can be obtained by backtracking along the steepest descent path from the goal. Therefore, this method provides an effective local-minima-free solution to the central problem in path planning, which is finding the shortest path between a starting and a goal location in a complex and dynamic environment.

Wavefront propagation methods are particularly well-suited for globally optimal one-to-all path planning in dynamic grid-based environments due to their local-minima-free nature. The wavefront can be efficiently recomputed to account for changes in the environment, such as moving obstacles or new obstacles being introduced. This makes wavefront propagation an attractive choice for applications such as autonomous vehicles, mobile robots, and drones, where real-time path planning is essential. Applications include scenarios where distances to all points in the grid are needed, such as coverage path planning, multi-robot path planning, and warehouse automation.


As powerful as marching methods currently are, they all share three major limitations. The purpose of this paper is to address those limitations with specific emphasis on making such methods more suitable for globally optimal one-to-all 2D path planning. 

\paragraph{First Limitation: Approximate Solutions Due to Discretization}
The first limitation is that these methods provide approximate --- not exact --- wavefront propagation solutions. They all rely on discretization techniques to approximate the solution of the Eikonal equation that governs the wavefront propagation. Additionally, discretization is used again to approximate the gradient of the propagated function to obtain the path. This introduces further errors that can accumulate and lead to sub-optimal paths.

\paragraph{Second Limitation: Inefficiencies in Pushing Optimality and Deriving the Paths}
The current state-of-the-art~\cite{cancela2021wmm} proposes a “super-resolution wave marching method SR-WMM” that uses advanced interpolation techniques to reduce the $L_{1}$-norm error to $6e-5$ against an analytic 2D isotropic Euclidean distance function in an empty $101\times101$ grid. 
The reported $L_{1}$-norm error can be only obtained using $\text{SR-WMM}_{gs}^{5}$, which is the most accurate solution that~\cite{cancela2021wmm} provides as an open-source code. $\text{SR-WMM}_{gs}^{5}$ requires around 30 minutes of compute time for a $768\times768$ complex environment on an Intel\textregistered~Core\texttrademark~i7--12800H Processor, signaling an accuracy-efficiency trade-off.
Moreover, current marching methods do not provide paths inherently. To obtain those, one would need to perform an additional gradient descent (GD) step per path query. This makes this class of methods unsuitable for dynamic environments as recomputing an additional GD step per path query online can be quite expensive.

\paragraph{Third Limitation: Performance Degradation in Non-Empty Environments}
The third limitation of marching methods is that their performance deteriorates considerably in non-empty environments. Errors in $L_{1}$, $L_{2}$, and $L_{\infty}$ norms accumulate as the environment becomes more cluttered. Typically, those methods provide benchmarks against analytic functions in empty environments. As such, there is no established way to measure their actual performance in obstacle-rich environments.

To address these limitations, we propose a wavefront propagation algorithm that has the following properties:

\begin{enumerate}
  \item Exact: it uses the analytical expression of distance functions directly in its evaluations instead of distance approximation techniques.
  \item Produces and outputs globally optimal paths to all points in the grid as part of the algorithm itself, thus eliminating the need for additional gradient descent steps per query.
  \item Verified against Anya, the de facto state-of-the-art optimal 2D any-angle path planner~\cite{haraborANYA, Uras2015AnEC}. We establish this verification as a standard to quantify the performance of state-of-the-art marching methods in non-empty environments. Based on our findings, the $L_{1}$, $L_{2}$ and $L_{\infty}$ norms provided by such methods deteriorate in obstacle-rich environments, whereas our method maintains optimality.
\end{enumerate}

Additionally, in this paper, we show that the proposed algorithm is linear in compute and memory with the size of the grid $n$ and runs faster than state-of-the-art marching algorithms. Lastly, we open source our code on GitHub for reproducibility \href{https://github.com/IbrahimSquared/visibility-based-marching}{https://github.com/IbrahimSquared/visibility-based-marching}.

This paper is organized as follows:
\begin{itemize}
  \item in Section~\ref{sec:related_work}, we provide an overview of related work.
  \item In Section~\ref{sec:VBM}, we provide an overview of the proposed algorithm's working principle. We discuss the properties, efficiency and complexity of both the underlying visibility algorithm and the visibility-based marching (VBM) algorithm.
  \item In Section~\ref{sec:results}, we discuss experimental results. Specifically, we report on computing and benchmarking marching results, marching multiple cost functions, and marching from multiple starting positions.
  \item In Section~\ref{sec:conclusion}, we discuss future work and offer concluding remarks. 

\end{itemize}

\section{Related Work}\label{sec:related_work}
Several approaches to tackle wavefront propagation in grids have been proposed. One of the earliest attempts to address this problem is Dijkstra's classical algorithm~\cite{dijkstra1959note}, which considers each grid cell as a vertex and its connections to nearby neighbors as edges. The algorithm explores the grid iteratively by propagating the distance between each visited vertex and its neighbors. The number of neighbors is typically limited to either four (i.e., north, south, east, and west) or eight (i.e., including the diagonal neighbors north-east, north-west, south-east, and south-west). To consider more neighbors, line-of-sight checks are typically required to determine if paths exist between the vertex and those neighbors. To achieve exact wavefront propagation using Dijkstra's algorithm, one would need to iterate over the entire grid from every vertex. This quickly becomes highly inefficient, asides from the impracticality associated with performing a line-of-sight check for every edge transition. 

A continuous Dijkstra type of algorithms that is not limited to a predefined set of neighbors was later set forth. In~\cite{mitchell1993SPM}, Mitchell introduced a continuous Dijkstra paradigm that has an $\mathcal{O}(V^{{\frac{5}{3}} + \epsilon})$ time and $\mathcal{O}(V\log{}V)$ space complexities, where $V$ is the number of obstacle vertices, and for any fixed $\epsilon > 0$. It also requires an additional $\mathcal{O}(\log{}V)$ compute per path query. The time and space complexities were both reduced by Hershberger and Suri~\cite{hershberger1999SPM} to $\mathcal{O}(V\log{}V)$. Nearly two decades later, Wang~\cite{wang2021SPM} managed to reduce space complexity to $\mathcal{O}(V)$. This class of methods, however, is not ubiquitous in literature due to its many limitations. Apart from scaling with the number of obstacle vertices, this approach is limited to polygonal domains (not applicable on grid maps) and does not perform well in 3D as dealing with obstacle vertices in 3D is impractical.

To address the limitations of previous wavefront propagation methods, Sethian introduced the fast marching method (FMM)~\cite{Sethian1996Marching}. This method entails advancing the solution of the Eikonal equation, a particular Hamilton-Jacobi non-linear first-order partial differential equation (PDE), in small steps along the independent variable, such as time or space. The Eikonal equation governs the monotonic evolution of wave frontlines and can be expressed as:
\begin{equation}
    1 = F(X) ||\nabla T(X)||,\; \; \; \; T(\Gamma_{0}) = 0,
    \label{eq:EikonalEquation}
\end{equation} 
where $F(X)$ and $T(X) \in \mathbb{R}$ are the expansion speed of the wave and its time-of-arrival respectively as a function of the position $X = (x_{1}, x_{2},\ldots, x_{d}) \in \mathbb{R}^{d}$. $\Gamma$ constitutes the boundary of the frontline, with $T(\Gamma_{0}) = 0$ being the initial boundary condition. A solution of~\eqref{eq:EikonalEquation} has been outlined in Osher and Sethian's previous work~\cite{OSHER1988LevelSets} based on discretizing the gradient and solving it via first-order accurate entropy-satisfying upwind schemes. 

In 2D, the original FMM algorithm begins at one or more starting grid positions, then solves a quadratic equation involving gradients for each of the four neighbors (north, south, east, west). The original starting positions are then marked as visited, and the neighbors that just got their quadratic equations solved are marked as the next starting positions. This process, although simplified here, is repeated iteratively until all of the grid cells are marked as visited. FMM, in turn, is limited to a cell's four direct neighbors. A method called multistencils fast marching method (MSFM) was later put forth by Hassouna and Farag~\cite{hassouna2007MSFM} that extends to all eight neighbors in 2D using two stencils and all 26 neighbors in 3D using six. Stencils are arrangements of nearby cells that are useful when solving PDEs. MSFM also introduces the usage of second-order accurate gradients. The authors showed that such a method retains a comparable solution complexity of $\mathcal{O}(n\log{}n)$ yet yields more accurate results than the original FMM, where $n = n_{x} \times n_{y}$ is the size of the grid.

Several other variations of FMM have also been introduced that either have the same performance~\cite{Zhao2004AFS}, a better  complexity of $\mathcal{O}(n)$~\cite{kim2001On,kao2005On}, or even a higher accuracy~\cite{BRUYNEEL201345, ahmed2011FMM, chopp2001fmm, covello2003fmm, cancela2015wmm, merino2018fmm, cancela2021wmm}.

Visibility between grid cells in grid-based environments is most commonly evaluated using ray-casting methods. One commonly adopted voxel traversal algorithm was presented by Amanatides et al.\cite{Amanatides1987Traversal}. Such algorithms are designed to evaluate visibility of one pixel at a time and are inefficient when it comes to evaluating visibility of an entire grid due to repetitive and redundant calculations. In 2024, Ibrahim et al.~\cite{ibrahim2024efficient} proposed an efficient method for evaluating holistic visibility for an entire grid that is based on a single-pass, dynamic-programming approach that runs up to two orders of magnitude faster than typical ray-casting. The method is based on a first-order upwind scheme solution of the advection equation, a linear first-order hyperbolic partial differential equation. The authors define visibility as a transportable quantity and transport it from a starting position to all other grid cells in a highly efficient manner. 

In order to showcase the effectiveness of the proposed visibility algorithm, the authors of~\cite{ibrahim2024efficient} also introduced a heuristic path planner that uses the quick visibility evaluations to explore the environment and connect the starting position to a goal position. This results in a one-to-one any-angle heuristic path planner that is suboptimal in both compute time and path length. Optimal one-to-one 2D path planners, however, have been around for quite some time. The most notable one is Anya~\cite{haraborANYA, Uras2015AnEC}, a state-of-the-art optimal 2D any-angle path planner capable of finding globally optimal paths in 2D grid-based environments in real-time. Other graph traversal methods, such as $A^\ast$\cite{hart1968formal}, are limited by their dependence on the graph structure. In typical grid-based path planning applications, $A^\ast$ only considers each node's eight direct neighbors. To obtain globally optimal any-angle paths, $A^\ast$ would need to consider a fully connected graph, where each node is connected to every other node in the grid. The worst-case time complexity of $A^\ast$ is $\mathcal{O}(b^{d})$, where $b$ is the branching factor and $d$ is the depth of the solution. As the number of neighbors $b$ increases, the computational complexity grows exponentially, making it impractical to obtain globally optimal paths.

\section{Visibility-based Marching}\label{sec:VBM}
The algorithm's working principle is simple and elegant and can be expressed in a single paragraph. 

Marching begins from one or more starting positions which are first inserted into a priority queue. At the initial step, all starting positions have a distance of zero, but otherwise, the priority queue ensures that marching is always done on a minimum-distance basis. The pixel position having the shortest distance is popped from the priority queue, and visibility between its parent node and each of the current position's eight neigbours is then queried. If a neighbour is visible to the designated parent node, the analytic distance to it is directly computed, and it is inserted into the priority queue. If the neighbour is not visible to the parent node, a pivot is created at the current position, and the distance to the neighbour is computed as the sum of the distance from the parent node to the pivot and the distance from the pivot to the neighbour. The newly created pivot can now potentially be a parent to subsequent pixels. The neighbour is then marked as updated. Information about potential parent nodes is propagated locally (saved on the go) along the wavefront in the form of a matrix $\mathcal{C}$ and conflicts between multiple potential parent nodes are resolved by picking the visible parent that results in the least overall distance. The algorithm continues until all grid cells are visited. Implementation details are highlighted in Alg.~\ref{alg:wavefront}.

\begin{algorithm}
\caption{Visibility-based Marching Algorithm}\label{alg:wavefront}
\begin{algorithmic}[1]
  \State{} Initialize priority queue $Q$ with starting positions
  \State{} Set initial distances of all starting positions to zero
\While{$Q$ is not empty}
    \State{}Pop pixel position $p$ with shortest distance from $Q$
    \For{each neighbor $n$ of $p$}
        \State{}Query visibility between $n$ and parent of $p$, $g$
        \If{$n$ is visible to $g$}
            \State{} Compute  distance from $g$ to $n$
        \Else{}
            \State{} Create pivot at $p$
            \State{} Compute distance from $g$ to $p$ and $p$ to $n$
        \EndIf{}
        \State{} Insert $n$ into $Q$ with updated cumulative distance
    \EndFor{}
    \State{}Mark $p$ as updated
  \EndWhile{}
  \State{}Internal Subroutines:
    \begin{itemize}
        \item Propagate information about potential parent nodes $g$ using matrix $\mathcal{C}$
        \item At every step, check nearby updated neighbors for potential parents
        \item Resolve conflicts by selecting out of the visible parents the one causing least overall distance
        \item Check any additional exit criterion
    \end{itemize}
\end{algorithmic}
\end{algorithm}

As such, the algorithm constructs shortest paths from one or more starting points to all other points in the grid by building a hierarchy of parent and child nodes. The hierarchy is saved in matrix $\mathcal{C}$, where each grid cell contains the parent node that results in the shortest path to it. In turn, each subsequent parent node points to its own parent, and so on, until the starting point is reached, which is marked as the parent of itself. The algorithm is not greedy in the sense that it computes the shortest path to all points in the grid from one or more starting positions rather than have a greedy guiding heuristic towards a single goal. The version that does the latter, dubbed $V^\ast$, is provided in the supplementary material and is further discussed in Section~\ref{sec:v_star}.

Visibility queries are evaluated using an underlying routine that is based on the method proposed in~\cite{ibrahim2024efficient}. Information regarding visibility --- which pivots see which cells --- is stored in a hashmap, which gets updated as the wavefront propagates.

The hashmap employed in~\cite{ibrahim2024efficient} has an amortized constant time complexity of $\mathcal{O}(1)$ for access, insertion, and deletion operations and is more optimized than std::unordered\_map from the C++ standard library. The authors show and experimentally verify that their proposed algorithm has a linear complexity with the number of cells due to its single-pass nature. Moreover, the priority queue employed in our code, std::priority\_queue, is a binary heap that has a logarithmic time complexity of $\mathcal{O}(\log{}n)$ for insertion and deletion operations. The overall theoretical time complexity of VBM is $\mathcal{O}(n\log{}n)$. However, for the range of grid sizes relevant in proposed applications, the bottleneck is pronouncedly in the hashmap. Hence we have an effective $\mathcal{O}(n)$ time complexity in this range, as shown in Section.~\ref{sec:results} where we also discuss the memory complexity of the proposed algorithm. $n$ in this context stands for the number of grid cells in the map.

\begin{figure}[H]
    \centering
    \includegraphics[width=8cm,keepaspectratio]{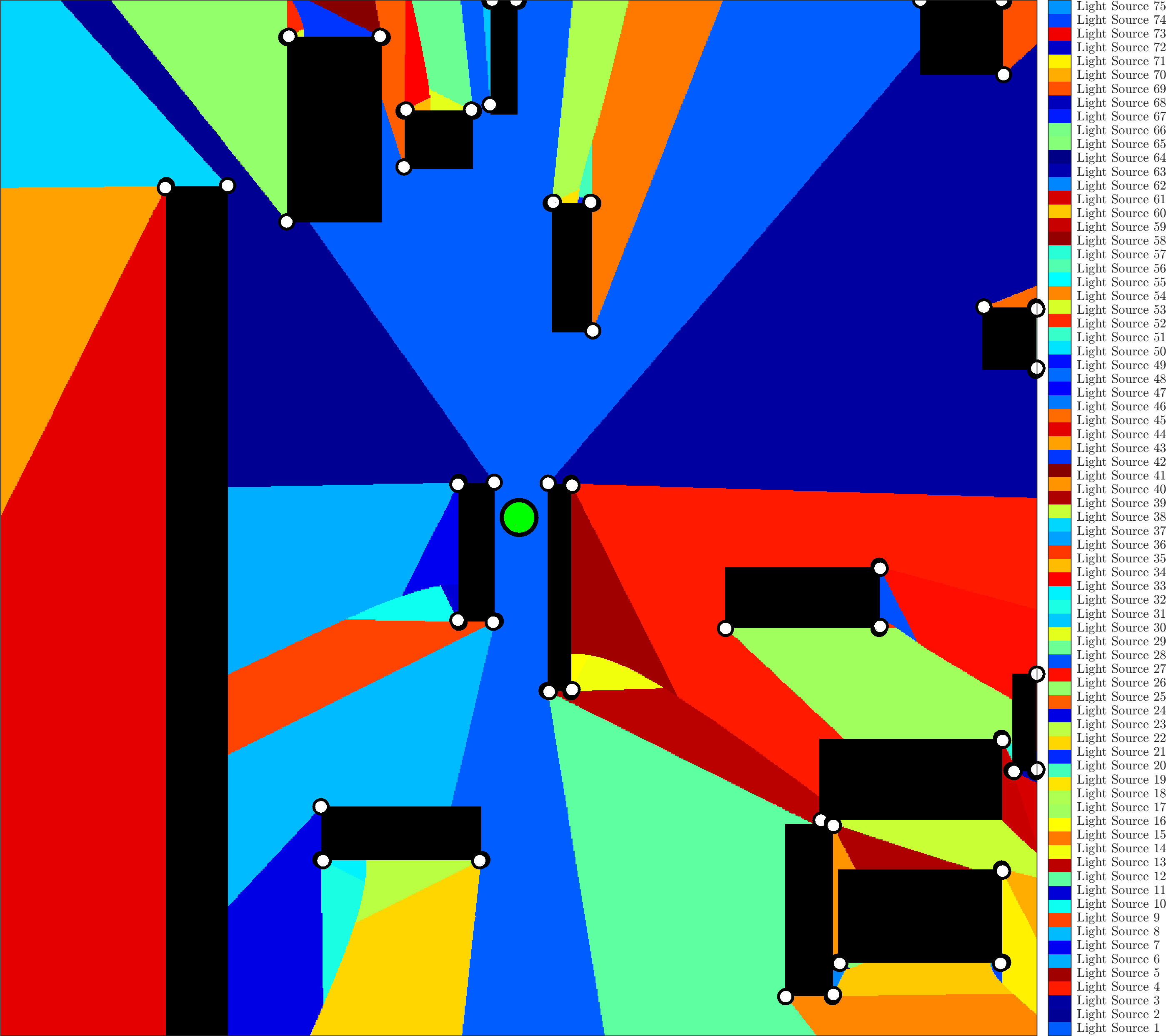}
    \caption{Resulting $\mathcal{C}$ matrix from VBM when applied to the same environment as in Fig.~\ref{fig:contours_1}. Original wave source is the green dot at the center, obstacles are shown in black, and the pivots are shown as white dots. Every unique color refers to grid cells that have the same parent node.}\label{fig:cameFrom}
\end{figure}

The result is a single-pass highly efficient algorithm that is capable of producing globally optimal paths to all points in the grid from one or more starting positions. A sample result of such a process when applied in an obstacle-rich environment is illustrated in Fig.~\ref{fig:cameFrom}. One output of VBM is the matrix $\mathcal{C}$ containing the parent node for every grid point. We illustrate in Fig.~\ref{fig:cameFrom} the matrix $\mathcal{C}$ that results from marching in the environment introduced in Fig.~\ref{fig:contours_1}. This matrix incorporates globally optimal paths from the origin to every position in the grid and is produced at no additional compute cost. Starting at any position in the grid, an agent can simply read the solution from this matrix and backtrack to reach the original source. All implementation details are provided in the supplementary material and the open-source code.

\section{Applications \& Experimental Results}\label{sec:results}
\subsection{Marching}
\begin{figure}[t]
    \centering
    \vspace{0.02cm}
    \includegraphics[width=7.5cm,keepaspectratio]{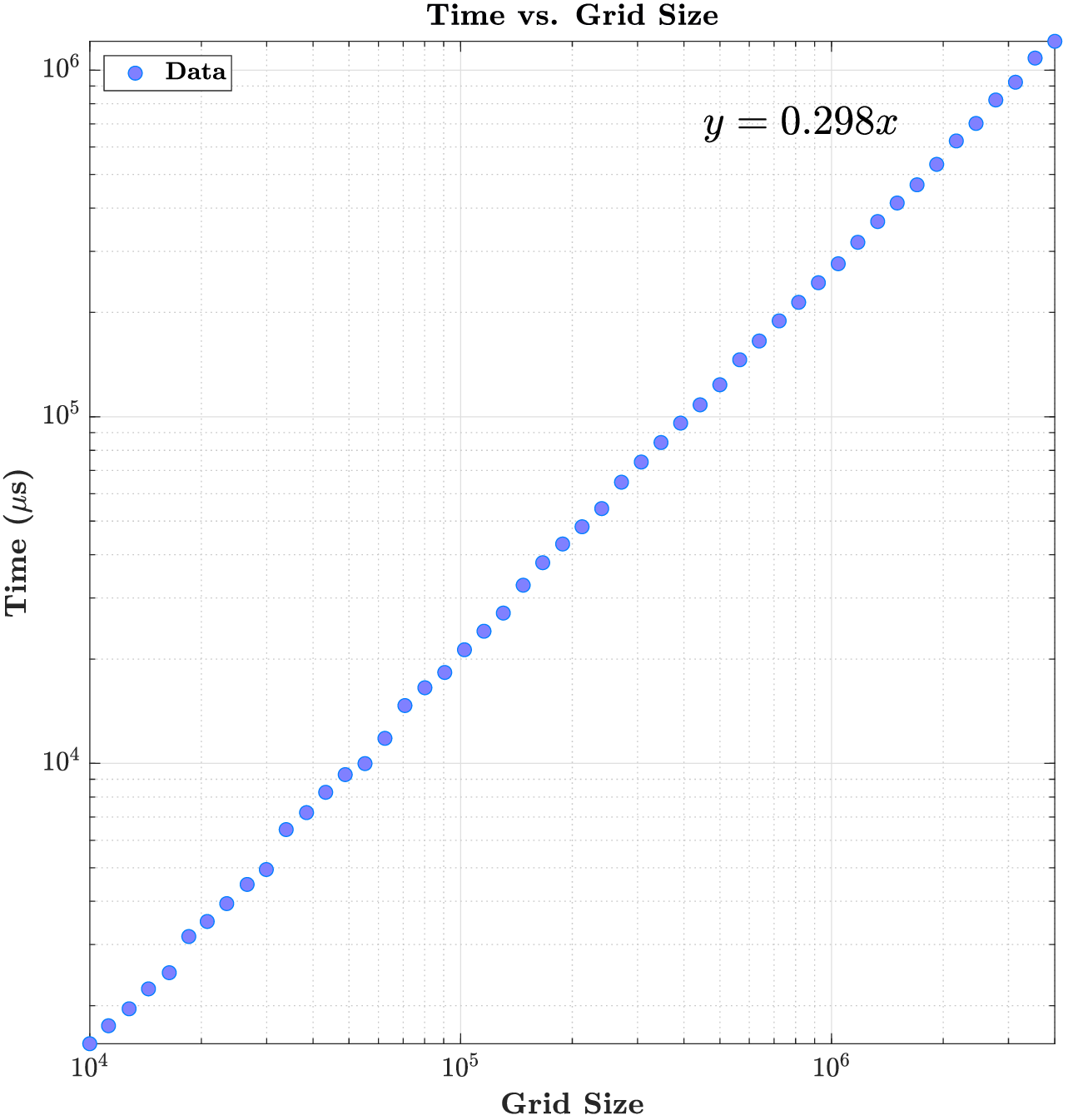}
    \caption{Log-log plot of average compute time of the proposed algorithm with respect to the grid size $n_{x}\times n_{y}$. The data is fit with linear regression. The resulting equation $y = 0.298x$ is shown on the plot and has an $R^{2}$ value of $0.998$ which signifies a strong confidence in the algorithm's linear complexity.}\label{fig:linear_complexity}
\end{figure}

VBM has a linear computational complexity with respect to the grid size $n_{x} \times n_{y}$. In order to demonstrate this, we carried out 50 sets of experiments, with each set comprising twenty repetitions, covering a range of logarithmically-sampled environment sizes. The range of sizes varied from $100 \times 100$ to $2000 \times 2000$. The experiments were conducted in C++ on an Intel\textregistered~Core\texttrademark~i7--9750H Processor. For the purpose of quantifying the complexity of our proposed algorithm, we intentionally chose an empty environment without any obstacles to provide a conservative estimate of compute time. In VBM, occupied cells are not traversed, so environments with more obstacles actually result in lower compute times due to fewer cells being processed. Consequently, the provided estimates in obstacle-free environments represent a worst-case scenario, with the algorithm performing linearly faster as fewer cells are visited. As illustrated in Fig.~\ref{fig:linear_complexity}, the resulting average compute time increases linearly with the number of grid cells at a slope of \SI{0.298}{\micro\second} per grid cell. Enabling marching through obstacles only minimally affects the slope of the line, while the overall linearity of the scheme is maintained.
Moreover, VBM's memory complexity is proportional to the grid size $n_{x}\times n_{y}$. All the used data structures contain at most $n_{x} \times n_{y}$ elements, except for the visibility hash map which contains at most $2 \times n_{x} \times n_{y}$ in obstacle-rich environments. This is due to the fact that a grid point may be seen by more than one parent node. 

In order to fully validate the performance of our algorithm, we march multiple cost functions in multiple scenarios:
\subsubsection{Single Origin \& Multiple Distance Functions}
\begin{figure}[t]
    \centering
    \subfigure[$L_{1}$ distance (Manhattan distance)] {
        \includegraphics[width=0.23\textwidth]{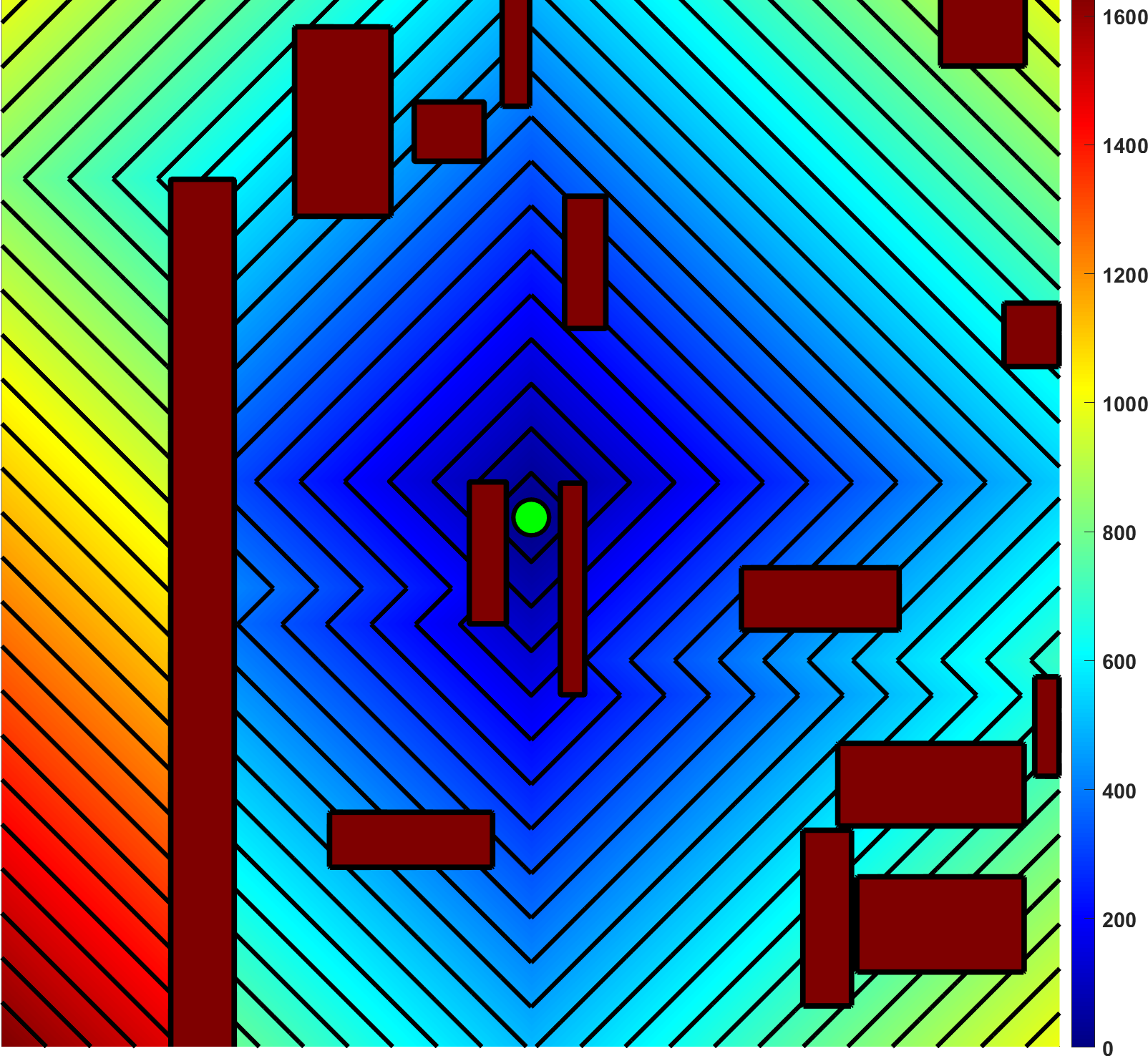}
    }\hspace*{\fill}
    \subfigure[$L_{3}$ distance (Cubic distance)] {
        \includegraphics[width=0.23\textwidth]{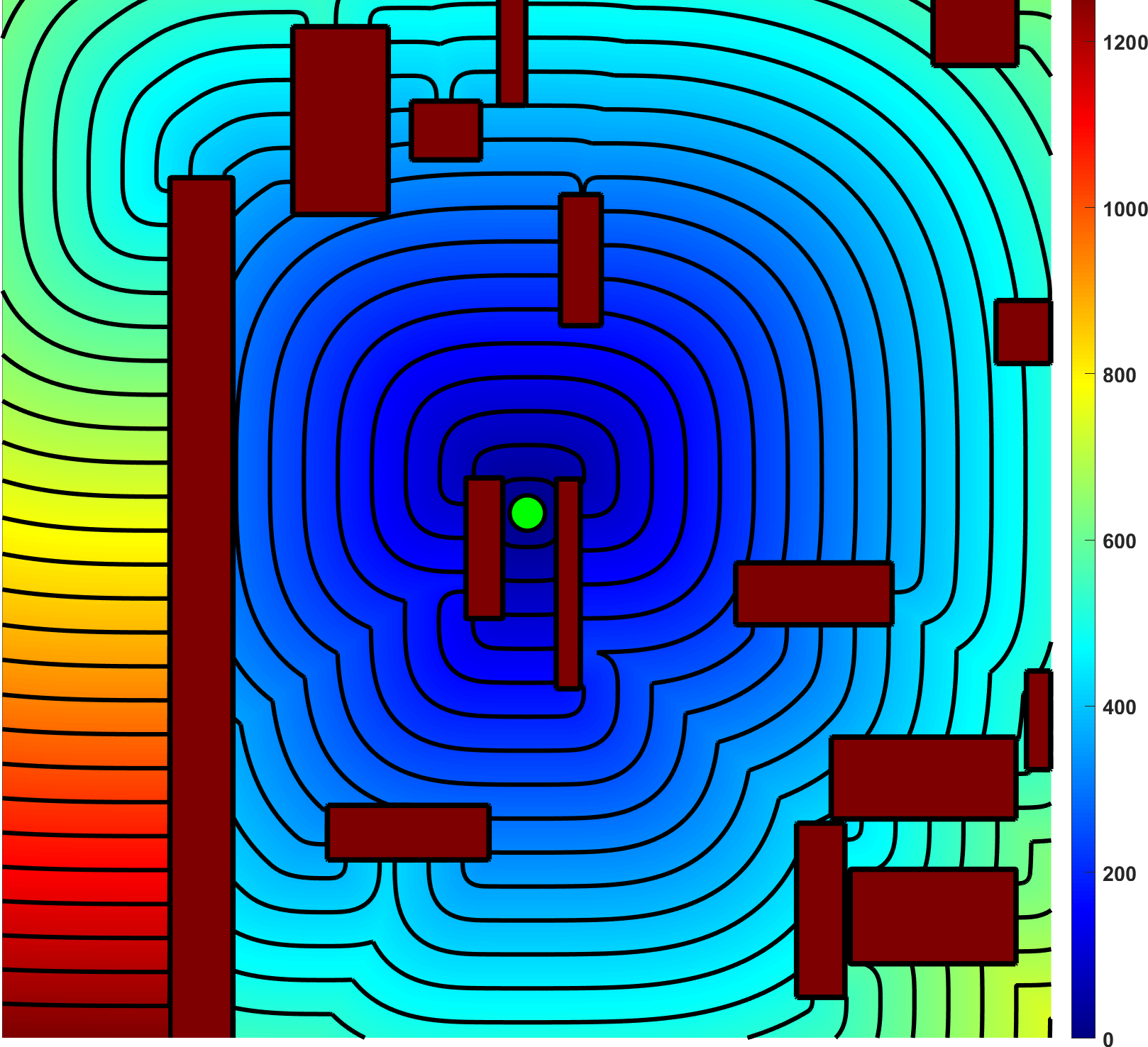}
    }
    \subfigure[$L_{\infty}$ distance (Chebyshev distance)] {
        \includegraphics[width=0.23\textwidth]{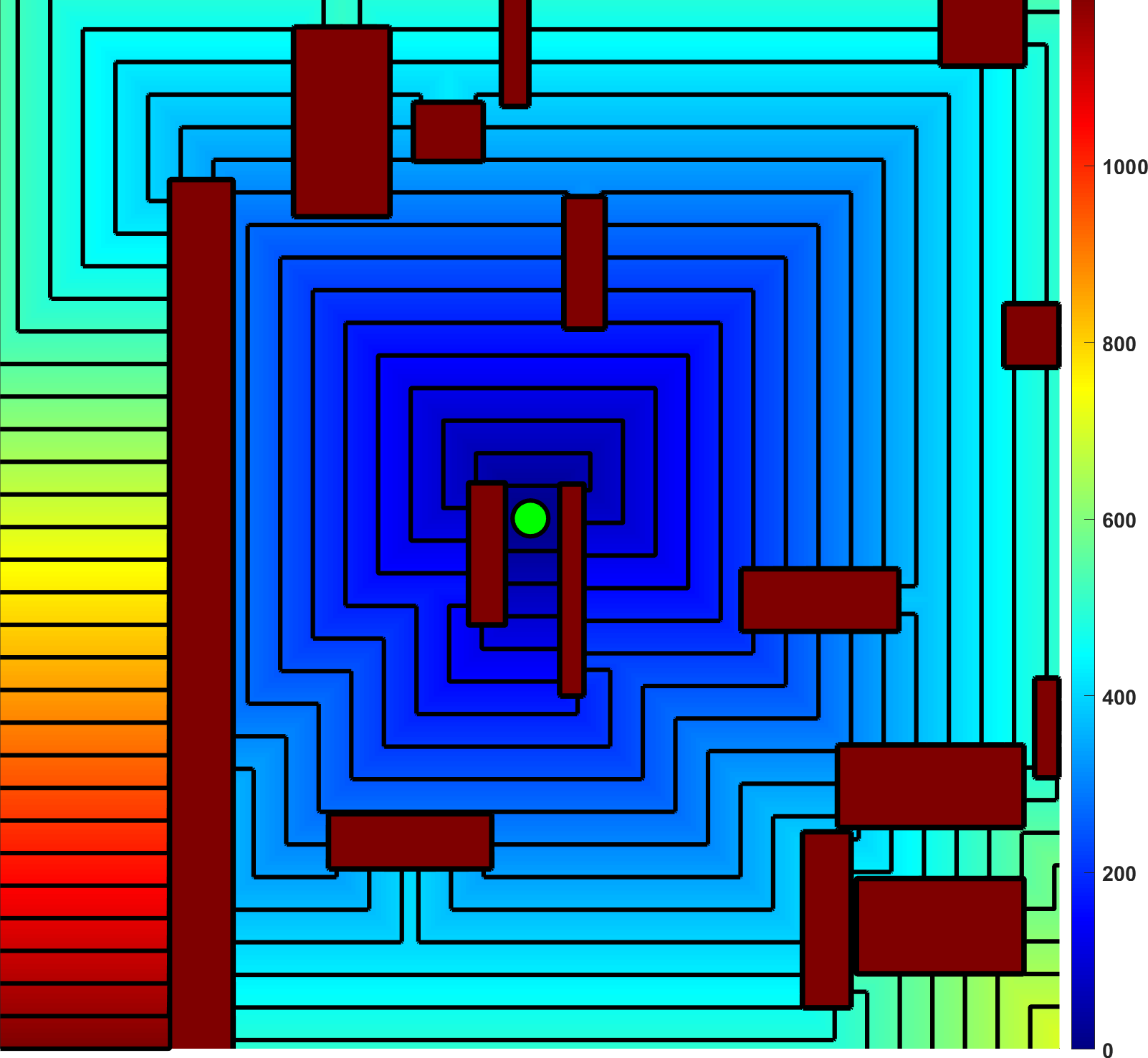}
    }\hspace*{\fill}
    \subfigure[Quasi-Euclidean distance] {
        \includegraphics[width=0.23\textwidth]{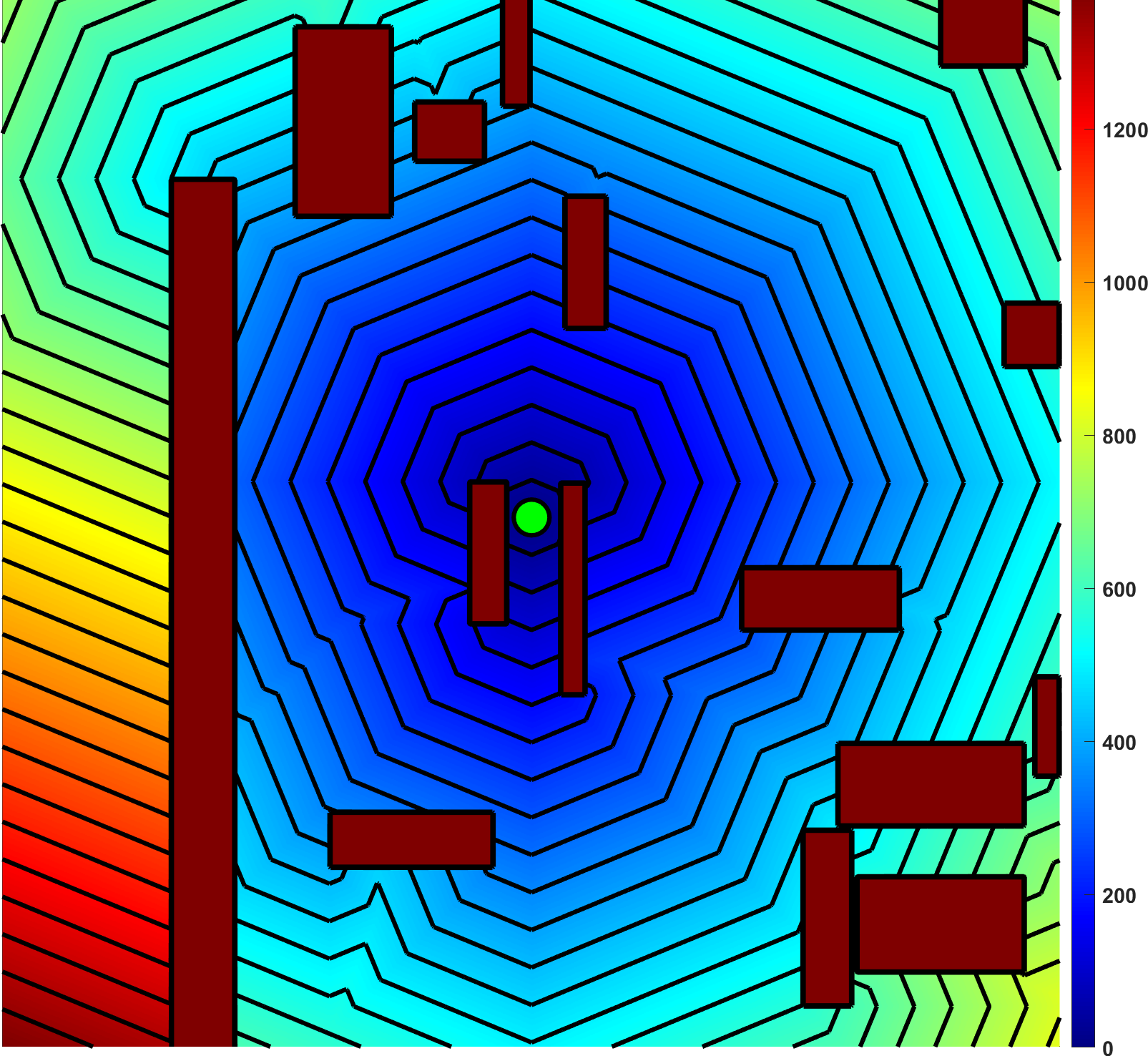}
    }
    \caption{Comparison of marching results when utilizing different distance metrics. (a) $L_{1}$ distance, (b) $L_{3}$ distance, (c) $L_{\infty}$ distance, and (d) Quasi-Euclidean distance. Fig.~\ref{fig:contours_1} shows the $L_{2}$ (Euclidean) distance. }\label{fig:multiple_distance_functions}
\end{figure}

The distance function that VBM propagates need not be Euclidean --- Minkowski distance with $p = 2$,
\begin{equation}
  D(X, Y) = \Big(\sum_{i=1}^{d} ||x_{i} - y_{i}||^{p}\Big){}^{\frac{1}{p}},
\end{equation}
where integer $p$ is the order of the Minkowski distance and $X = (x_{1}, x_{2},\ldots,x_{d})$ and $Y = (y_{1}, y_{2},\ldots,y_{d}) \in \mathbb{R}^{d}$. As illustrated in Fig.~\ref{fig:multiple_distance_functions}, we demonstrate the propagation of Manhattan ($p = 1$), Cubic ($p = 3$), Chebyshev ($p = \infty$), and quasi-Euclidean distance functions. This goes to show that VBM is not limited to Euclidean distance functions and can be used with any Minkowski distance function. The choice of which distance function to use is application-dependent. For example, the Chebyshev distance is suitable for environments with diagonal movements whereas the Cubic distance is suitable for environments where the cost of movement is proportional to the cube of the distance. Euclidean distance is most commonly used in practice due to its geometric properties and the fact that it is the most natural and intuitive.

\subsubsection{Multiple Origins \& Single Distance Function}
\begin{figure}[ht]
    \centering
    \vspace{0.1cm}
    \includegraphics[width=8cm,keepaspectratio]{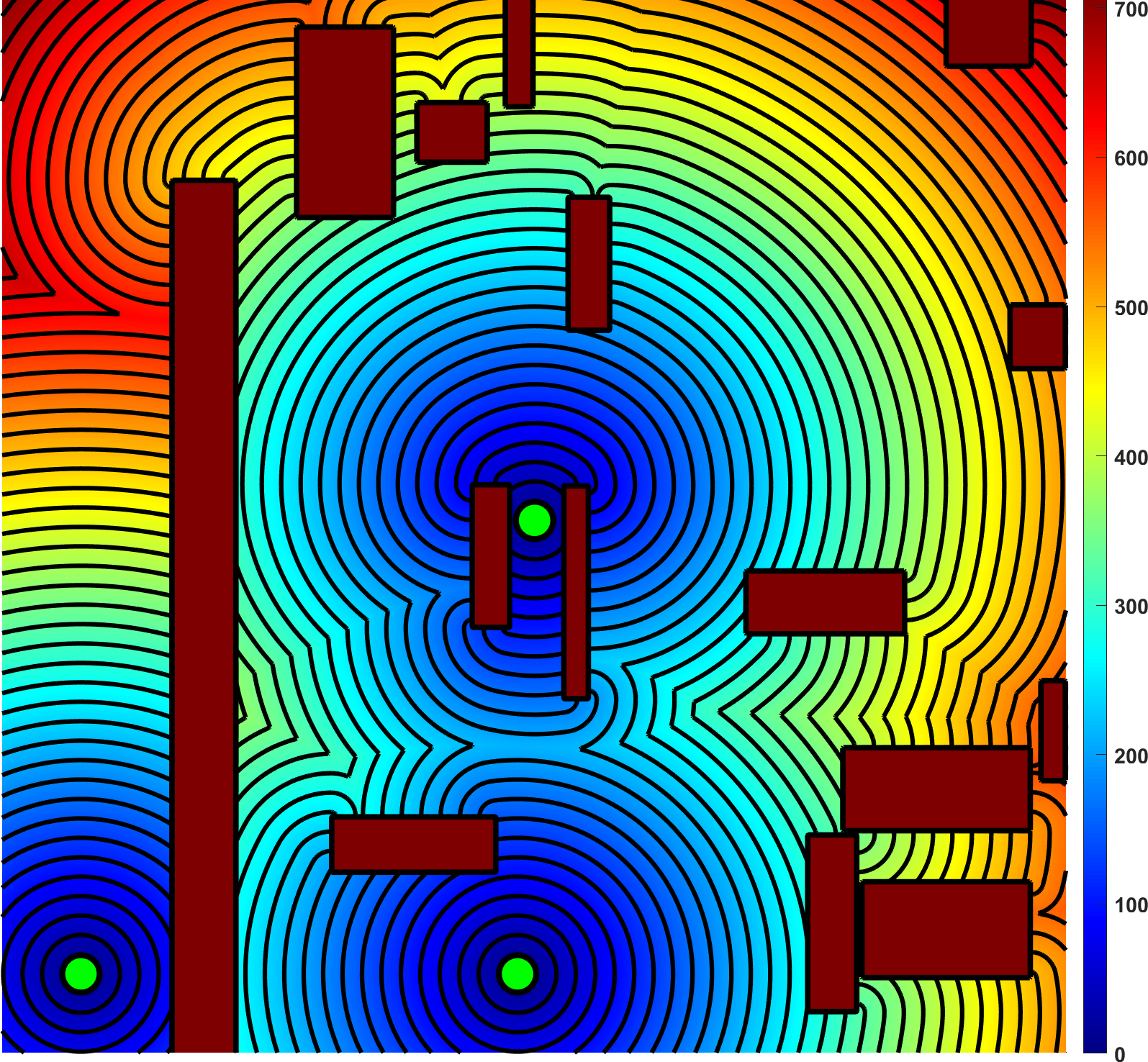}
    \caption{Multiple sources for marching to begin from marked as green dots. Same environment as in Fig.~\ref{fig:contours_1}. No collisions or numerical issues arise.}\label{fig:multiple_sources}
\end{figure}

VBM is capable of propagating multiple wavefronts without numerical issues or wave collisions as illustrated in Fig.~\ref{fig:multiple_sources}. This could be useful in applications with multiple agents or multiple objectives. For example, a warehouse robot could use this capability to keep track of the shortest path to the nearest charging station, the nearest exit, and the nearest item to pick up. Another example is pursuit-evasion games where an agent has to be aware of the distance to the closest pursuer and update its plans accordingly.

\subsubsection{Benchmarks with Anya, $\text{SR-WMM}_{gs}^{5}$, and MSFM}
\begin{figure}[ht]
    \centering
    \vspace{0.1cm}
    \includegraphics[width=8cm,keepaspectratio]{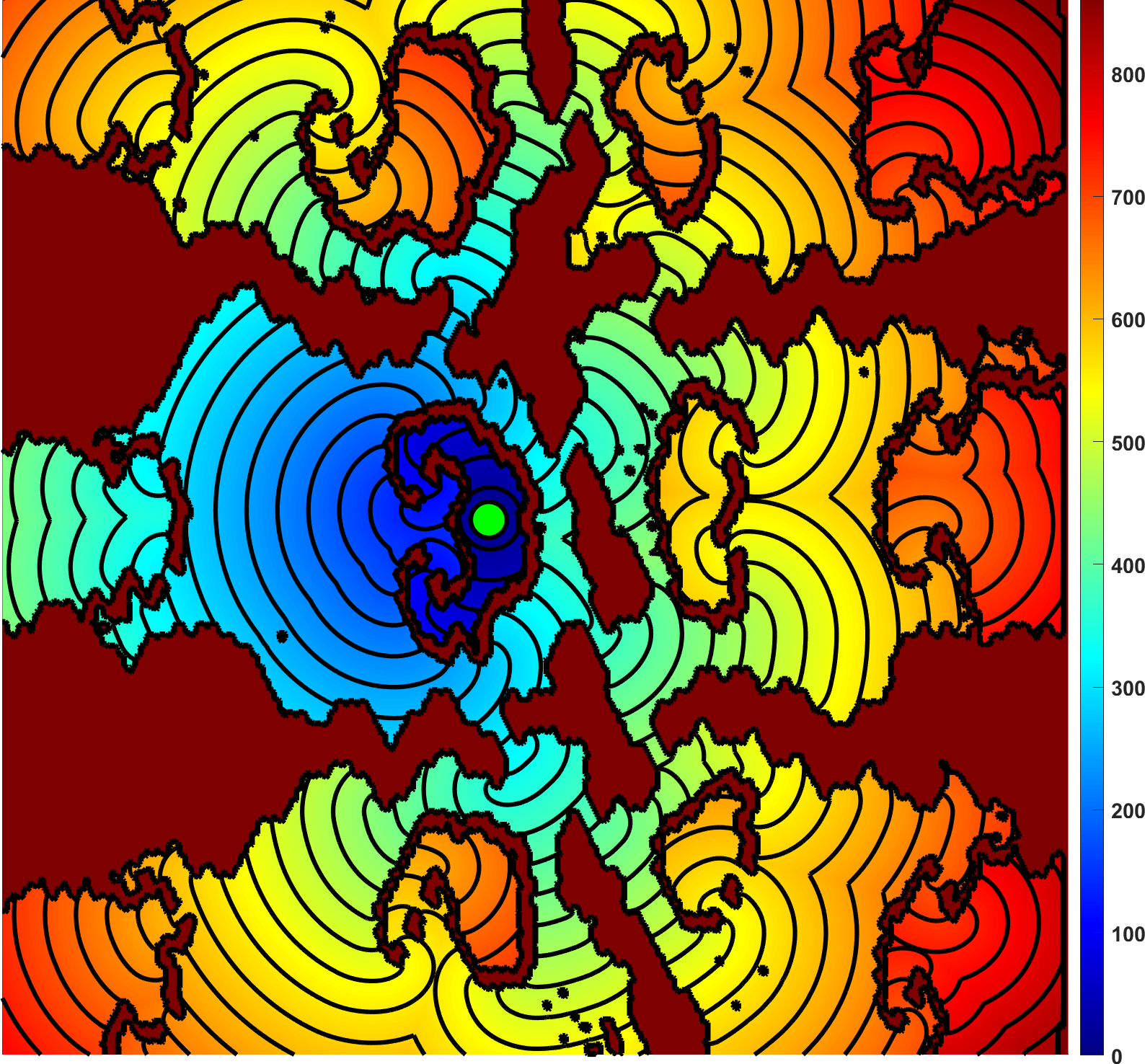}
    \caption{VBM applied to “AcrosstheCape”, a $768\times 768$ map provided along with Anya's source code that is often used in bench-marking path planners. Origin of the wave is marked at the center with a green dot.}\label{fig:AcrossTheCape_VBM}
\end{figure}

{\label{sec:benchmarking}}
Typically, state-of-the-art marching methods provide tables in which they compare the performance of their methods against other marching methods in empty environments. VBM naively beats all those benchmarks as it uses the analytic function in its evaluations and results in zero error. Instead, we use distances computed by Anya, a third-party reliable and optimal state-of-the-art 2D path planner~\cite{haraborANYA, Uras2015AnEC}, as a baseline for comparing propagated distance values computed by VBM in a sufficiently complex and non-empty environment, Fig.~\ref{fig:AcrossTheCape_VBM}, against those computed by other marching methods. Distances produced by Anya, however, only serve as a theoretical unattainable lower bound, since Anya makes the assumption that an agent has a radius of zero, allowing it to travel along edges of obstacles. Such an assumption may be suitable for computer games, but it is invalid in realistic environments and inapplicable to marching methods. In our case, the underlying visibility algorithm necessitates that at no point can the line-of-sight violate any occupied region when using a visibility threshold of $0.5$. 

\begin{figure}[ht]
    \centering
    \vspace{0.1cm}
    \begin{subfigure}{}
        \includegraphics[width=8.5cm]{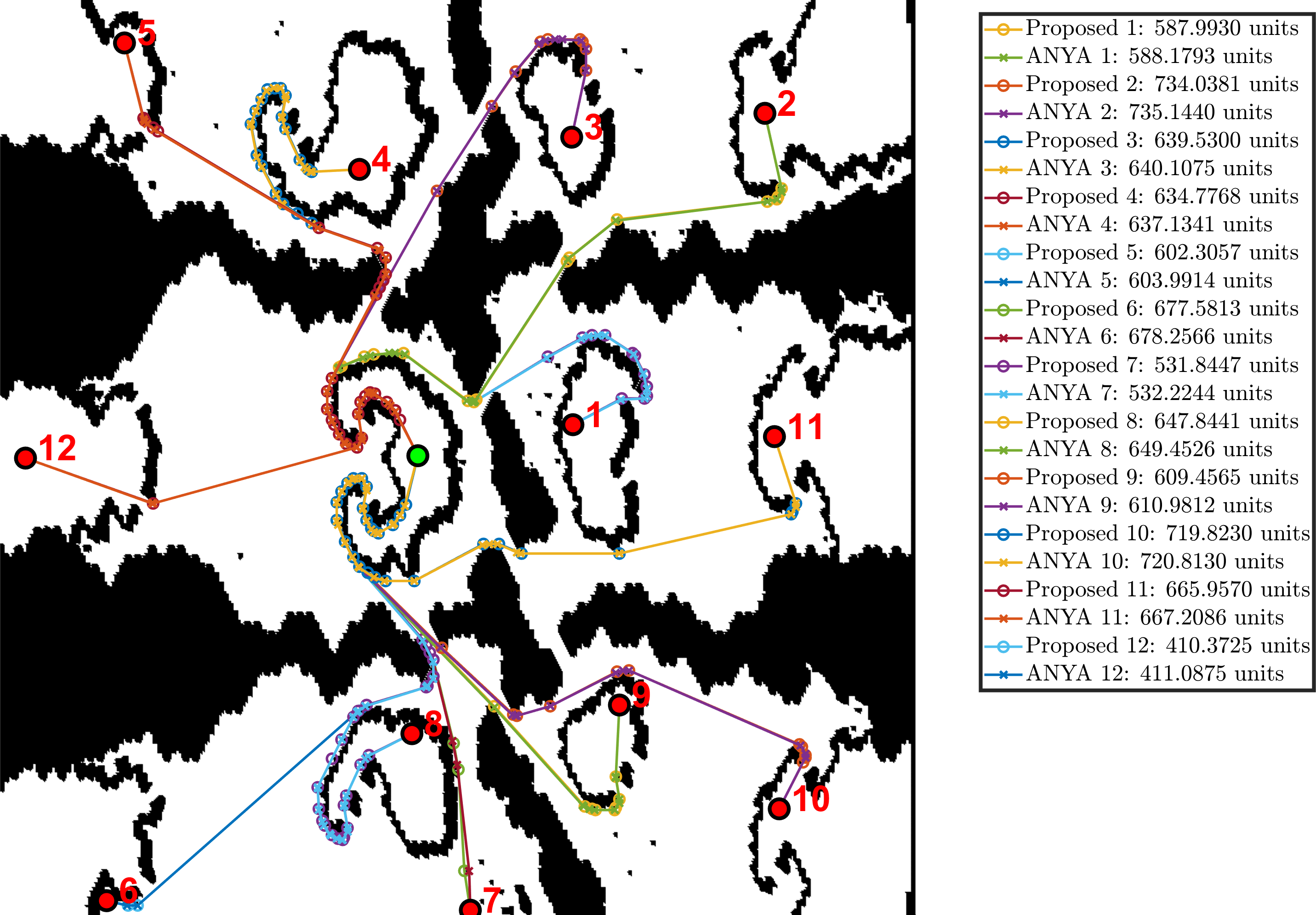}
    \end{subfigure}\hfil
    \caption{The paths starting from the central green dot and ending in red dots that were used to perform the comparison reported in Table.~\ref{tab:num_results}. The environment is the same as in Fig.~\ref{fig:AcrossTheCape_VBM}, but shown in gray-scale instead (obstacles are shown in black). Only VBM and Anya can provide globally optimal paths --- $\text{SR-WMM}_{gs}^{5}$ \& MSFM can only report distances.}\label{fig:Anya_before_after}
\end{figure}

As such, we compare against the current state-of-the-art marching method $\text{SR-WMM}_{gs}^{5}$ as well as against MSFM\@. For $\text{SR-WMM}_{gs}^{5}$, we directly use the source code provided by its authors, whereas for MSFM, we use the efficient implementation provided in~\cite{kroon2023accurate}. We also directly use the source code provided by the authors of Anya. We consider the paths illustrated in Fig.~\ref{fig:Anya_before_after} and report the obtained distance results in Table.~\ref{tab:num_results}.

From Table.~\ref{tab:num_results}, we observe that the error norms reported by previous state-of-the-art marching methods in empty environments deteriorate considerably in complex and non-empty environments. It is also noticeable that the percent error using $\text{SR-WMM}_{gs}^{5}$ is higher than the one obtained using older methods, namely MSFM\@. Moreover, it can be observed that VBM outperforms both methods by a large margin. As mentioned before, Anya-computed paths are unattainable due to Anya's assumptions on path traversability. For the sake of completion, we assign occupied cells that Anya-computed paths go through as unoccupied in the map then recompute using VBM, $\text{SR-WMM}_{gs}^{5}$, and MSFM\@. The results are reported both in Fig.~\ref{fig:Anya_before_after} and Table.~\ref{tab:num_results} as \emph{Anya-Compatible} paths. In conclusion, VBM outperforms state-of-the-art methods in both accuracy and efficiency when reliable third-party estimates are provided by Anya. 

\subsection{$V^\ast$ --- A Greedy Optimal Version of VBM}\label{sec:v_star}
A greedy optimal version of our algorithm, which we call $V^\ast$  is also provided in the supplementary material. It outperforms regular greedy planners such as $A^\ast$ in terms of shortest path length at comparable computational time in C++. The difference is that $V^\ast$ requires a lot less iterations to converge to the optimal path since each cell is visited at most once, whereas in $A^\ast$ each cell may be visited multiple times. $V^{\ast}$, however, deals with visibility evaluations, whose computational overhead balances out with the faster convergence of $V^\ast$. $V^\ast$ differs from VBM in that it is greedy and only computes the shortest path to a single goal as it is guided by a greedy heuristic towards the goal. Its termination criterion is establishing the path with the target, whereas the termination criterion of VBM is visiting all cells in the grid --- empty $Q$. $V^\ast$ is based on the same principles as VBM, but it is more suitable for applications where a single goal is known and the shortest path to it is desired rather than requiring paths and distance evaluations for an entire grid. Compared to VBM, it is more efficient in terms of compute time, but is less versatile as it is limited to a single goal. Since it runs the same underlying algorithm, it also produces globally optimal paths. It optimally propagates the heuristic distance function compared to $A^\ast$ because it is not limited to a predefined set of neighbors, thus overcoming $A^\ast$'s limitation. $V^\ast$ produces results that are comparable to those of Anya as reported in Table.~\ref{tab:num_results} and discussed in Section~\ref{sec:benchmarking}.

\begin{table*}[htbp]
    \centering
    \begin{threeparttable}
    \caption{Numerical Results of 12 Path Distances in “AcrosstheCape” Map (Fig.~\ref{fig:Anya_before_after}) Computed Using VBM (proposed) Compared To $\text{SR-WMM}_{gs}^{5}$ \& MSFM, Using an Intel\textregistered~Core\texttrademark~i7--9750H Processor.}\label{tab:num_results}
    \begin{tabular}{cccccccc}
        \toprule
        \multicolumn{1}{c}{\textbf{Path}} & \multicolumn{2}{c}{\textbf{VBM (Proposed)}} & \multicolumn{2}{c}{\textbf{$\text{SR-WMM}_{gs}^{5}$}} & \multicolumn{2}{c}{\textbf{MSFM}} & \multicolumn{1}{c}{\textbf{Ground Truth (Anya)}} \\

        \cmidrule(lr){2-3} \cmidrule(lr){4-5} \cmidrule(lr){6-7} 
         & \textbf{Regular} & \textbf{Anya-Compatible} & \textbf{Regular} & \textbf{Anya-Compatible} & \textbf{Regular} & \textbf{Anya-Compatible} & \\
         
         \midrule
         
          1 & 592.139 & 587.993 & 596.828 & 592.748 & 595.439 & 591.220 & 588.179  \\
          2 & 737.682 & 734.038 & 742.164 & 738.670 & 740.279 & 736.897 & 735.144  \\
          3 & 643.039 & 639.530 & 646.165 & 642.735 & 644.952 & 641.550 & 640.108  \\
          4 & 639.803 & 634.777 & 645.314 & 640.058 & 644.042 & 639.014 & 637.134  \\
          5 & 605.552 & 602.306 & 608.999 & 605.511 & 608.300 & 605.018 & 603.991  \\
          6 & 680.470 & 677.581 & 685.592 & 682.770 & 684.731 & 681.883 & 678.257  \\
          7 & 533.799 & 531.862 & 537.970 & 536.072 & 537.740 & 535.723 & 532.224  \\
          8 & 651.529 & 647.844 & 658.391 & 654.963 & 657.660 & 654.151 & 649.453  \\
          9 & 612.170 & 609.457 & 617.119 & 614.569 & 616.103 & 613.536 & 610.981  \\
          10 & 723.492 & 719.823 & 728.669 & 725.062 & 727.509 & 724.104 & 720.813  \\
          11 & 669.121 & 665.957 & 674.558 & 671.464 & 673.705 & 670.575 & 667.209  \\
          12 & 412.719 & 410.373 & 414.129 & 411.761 & 413.274 & 411.028 & 411.088  \\
          \midrule
          \textbf{Average Distance (pixels)} & 625.126 & 621.794 & 629.658 & 626.365 & 628.644 & 625.352 & 622.882 \\
          \textbf{Percent Error (\%)\tnote{1}} & \textbf{0.360} & \textbf{-0.174} & 1.088 & 0.559 & 0.925 & 0.397 & 0.000 \\
          \midrule
          \textbf{Compute Time (s)} & \multicolumn{2}{c}{\textbf{0.195}} & \multicolumn{2}{c}{1789} & \multicolumn{2}{c}{0.268} & --- \\
        \bottomrule
    \end{tabular}
    \begin{tablenotes}
    \item[1] Note: A negative percent error indicates that the computed distance is shorter than the one obtained with Anya.
    \end{tablenotes}
    \end{threeparttable}
\end{table*}

\section{Conclusion \& Future Work}\label{sec:conclusion}

In this paper, we presented an efficient exact wavefront marching algorithm that outperforms state-of-the-art methods in both accuracy and speed. The proposed method computes optimal paths from one or more starting positions in the gridmap to all other grid cells. Our proposed method relies on an underlying visibility evaluation routine that efficiently processes visibility queries using an upwind scheme solution to an advection PDE\@. We introduced a standard to benchmark the performance of marching methods in obstacle-rich environments. We demonstrate that the performance of state-of-the-art marching methods deteriorates while our method maintains optimality. Numerical experiments verify the validity and applicability of our approach, making it a powerful tool to use when dealing with distance functions and optimal paths, which are not only ubiquitous, but also often necessary in robot applications. Lastly, a greedy version of our method can be adopted as well --- $V^\ast$ --- that is optimal, outperforms $A^\ast$ in terms of number of iterations and optimality, and runs in real-time at comparable computational speed in C++ with respect to $A^\ast$. We provide open-source code for further implementation details and reproducibility.

For future work, we plan to extend the algorithm to 3D, as the underlying visibility algorithm is based on a scheme that extends to $\mathbb{R}^{d}$. The underlying visibility algorithm also extends to quantifying curves-of-sight, meaning it may be possible to account for robot/vehicle kinematic models within the marching process itself, i.e., perform kinematically-constrained marching. The latter can be Reeds-Shepp or Dubin vehicles whose distance functions are non-Minkowski by nature. An investigation on the applicability and use-case of such a method in non-uniform grids is also of interest. An initial investigation on using parallelization techniques or a vector/matrix-based solution to marching rather than doing serial marching also looks promising.

\section*{Acknowledgments}
The authors would like to thank Assistant Professor Florian Feppon for providing interesting discussions, Dr.~Daniel Harabor for clarifications about the optimal path planner Anya, and Associate Professor Brais Cancela for clarifications about the open-source code on wave marching methods. The authors acknowledge that a preliminary non-peer-reviewed presentation of this work was given at the 42nd Benelux Meeting on Systems and Control, Elspeet, The Netherlands, March 2023. The presentation was linked to a one-page abstract and did not include detailed methodologies or results. 
 
\bibliographystyle{bibtex/IEEEtran} 
\bibliography{bibtex/IEEEabrv, bibtex/bibliography}

\end{document}